\documentclass{opt2022} 

\usepackage[utf8]{inputenc} 
\usepackage[T1]{fontenc}    
\usepackage{hyperref}       
\usepackage{url}            
\usepackage{booktabs}       
\usepackage{amsfonts}       
\usepackage{nicefrac}       
\usepackage{microtype}      
\usepackage{xcolor}         
\usepackage{makecell, multirow}
\usepackage{algorithm}
\usepackage{algpseudocode}
\usepackage{wrapfig}
\usepackage{sidecap}
\usepackage{float}
\usepackage{sidecap}

\usepackage{amsmath,amsfonts,bm}
\usepackage{ amssymb }

\def\eqref#1{equation~\ref{#1}}

\def\1{\bm{1}}

\DeclareMathAlphabet{\mathsfit}{\encodingdefault}{\sfdefault}{m}{sl}
\SetMathAlphabet{\mathsfit}{bold}{\encodingdefault}{\sfdefault}{bx}{n}

\newcommand{\norm}[2]{\left\| #1 \right\|_{#2}}
\newcommand{\abs}[1]{\left| #1 \right|}

\usepackage{mathtools}

\usepackage{graphicx,amsfonts,amscd,amssymb,bm,url,color,latexsym,bbm,amsmath}
\usepackage{physics}
\allowdisplaybreaks
\usepackage[capitalize,nameinlink]{cleveref}

\renewcommand{\mathbf}{\boldsymbol}

\newcommand{\mb}{\mathbf}
\newcommand{\mc}{\mathcal}

\newcommand{\bb}{\mathbb}

\newcommand{\reals}{\bb R}

\newcommand{\eps}{\varepsilon}
\newcommand{\R}{\reals}

\newcommand{\paren}{\pqty}

\DeclareMathOperator{\conv}{conv}

\DeclareMathOperator{\sign}{sign}

\DeclareMathOperator{\st}{s.t.}

\newcommand{\wh}{\widehat}

\newcommand{\pygranso}{\texttt{PyGRANSO}}

\title[Optimization for Robustness Evaluation beyond $\ell_p$ Metrics]{Optimization for Robustness Evaluation beyond $\ell_p$ Metrics}

\optauthor{%
\Name{Hengyue Liang$^1$} \Email{liang656@umn.edu}\\
\Name{Buyun Liang$^1$} \Email{liang664@umn.edu}\\
\Name{Ying Cui$^1$} \Email{yingcui@umn.edu }\\
\Name{Tim Mitchell$^2$} \Email{tmitchell@qc.cuny.edu}\\
\Name{Ju Sun$^1$} \Email{jusun@umn.edu }\\
\addr $^1$University of Minnesota, Minneapolis, USA\\ 
$^2$Queens College of the City University of New York, New York City, USA}

\begin{document}

\maketitle

\vspace{-2em}
\begin{abstract}
Empirical evaluation of deep learning models against adversarial attacks entails solving nontrivial constrained optimization problems. Popular algorithms for solving these constrained problems rely on projected gradient descent (PGD) and require careful tuning of multiple hyperparameters. Moreover, PGD can only handle $\ell_1$, $\ell_2$, and $\ell_\infty$ attack models due to the use of analytical projectors. In this paper, we introduce a novel algorithmic framework that blends a general-purpose constrained-optimization solver \pygranso, \textbf{W}ith \textbf{C}onstraint-\textbf{F}olding (PWCF), to add reliability and generality to robustness evaluation. PWCF 1) finds good-quality solutions without the need of delicate hyperparameter tuning, and 2) can handle general attack models, e.g., general $\ell_p$ ($p > 0$) and perceptual attacks, which are inaccessible to PGD-based algorithms. 

\end{abstract}


\section{Introduction}\label{Sec:introduction}
In visual recognition, deep neural networks (DNNs) are not robust against perturbations that are easily discounted by human perception---either adversarially constructed or naturally occurring~\cite{szegedy2013intriguing,GoodfellowEtAl2015Explaining,hendrycks2018benchmarking,EngstromEtAl2019Exploring,xiao2018spatially,WongEtAl2019Wasserstein,LaidlawFeizi2019Functional,HosseiniPoovendran2018Semantic,BhattadEtAl2019Big}. A popular way of finding an adversarial perturbation (a.k.a adversarial attack) is by solving the \emph{adversarial loss} formulation~\cite{madry2017towards}: 
\begin{align} 
\label{eq:robust_loss}
\max_{\mb x'} \ell\paren{\mb y, f_{\mb \theta}(\mb x')} ~ \text{,} \quad
\st  \; \mb x' \in \Delta(\mb x) = \{\mb x' \in [0, 1]^n: d\paren{\mb x, \mb x'} \le \eps\}
\end{align}
Here, $f_{\theta}$ is the DNN model, and $\Delta(\mb x)$ 
is an allowable perturbation set with radius $\eps$ as measured by the metric $d$. 
Early works assume $\Delta(\mb x)$ is the $\ell_p$ norm ball intersected with the natural image box, i.e., $\{\mb x' \in [0, 1]^n: \norm{\mb x- \mb x'}_p \le \eps\}$, where $p= 1, 2, \infty$ are popular choices~\cite{madry2017towards,GoodfellowEtAl2015Explaining}. To capture visually realistic perturbations, recent works have also modeled nontrivial transformations using non-$\ell_p$ metrics~\cite{hendrycks2018benchmarking,EngstromEtAl2019Exploring,xiao2018spatially,WongEtAl2019Wasserstein,LaidlawFeizi2019Functional,HosseiniPoovendran2018Semantic,BhattadEtAl2019Big,laidlaw2021perceptual}. As for empirical robustness evaluation (RE), solutions of \cref{eq:robust_loss} lead to the worst-case perturbations to fool $f_{\mb \theta}$. 


But solving \cref{eq:robust_loss} is not easy: the objective is non-concave for typical choices of loss $\ell$ and model $f_{\mb \theta}$; for non-$\ell_p$ metrics, $\Delta(\mb x)$ is often a complicated nonconvex set. In practice, there are two major lines of algorithms: \textbf{(a) direct numerical maximization} that takes differentiable $\ell$ and $f_{\mb \theta}$, and tries direct maximization, e.g., using gradient-based methods~\cite{madry2017towards,croce2020reliable}. This often only produces a suboptimal solution and can lead to overoptimistic RE; 
\textbf{(b) upper-bound maximization} that constructs tractable upper bounds for the margin loss $\ell_{\mathrm{ML}} = \max_{i \ne y} f_{\mb \theta}^i (\mb x') - f_{\mb \theta}^y (\mb x')$, where $y$ is the true class of $\mb x$ 
, and then optimizes against the upper bounds~\cite{SinghEtAl2018Fast}. Improving the tightness of the upper bounding while maintaining tractability is still an active area of research.


Another formalism of robustness is the \emph{robustness radius} (or minimum distortion radius), defined as the minimal level of perturbation that causes $f_{\mb \theta}$ to change its predicted class:
\begin{align}  
\label{eq:min_distort}
    \min_{\mb x' \in [0, 1]^n}\;  d\paren{\mb x, \mb x'}  \quad \st \; \max_{i \ne y} f_{\mb \theta}^i (\mb x') \ge f_{\mb \theta}^y (\mb x') 
\end{align} 
Solving \cref{eq:min_distort} produces not only a minimally distorted perturbation $\mb x'$, but also a robustness radius, which makes it another popular choice for RE~\cite{CroceHein2020Minimally, croce2020reliable, PintorEtAl2021Fast}. In fact, \cite{CroceHein2020Minimally, ZhangEtAl2021Revisiting, PintorEtAl2021Fast} perform adversarial attacks by trying to solve \cref{eq:min_distort}.

In this paper, we focus on numerical optimization of \cref{eq:robust_loss}. In particular, we \textbf{(I)} adapt the constrained-optimization solver \texttt{PyGRANSO} ~\cite{curtis2017bfgs,BuyunLiangSun2021NCVX} with a constraint-folding (PWCF) technique---crucial for making \texttt{PyGRANSO} solve \cref{eq:robust_loss} with reasonable speed and quality, and \textbf{(II)} show that PWCF can handle \emph{attacks other than the $\ell_1$, $\ell_2$, and $\ell_\infty$ ones}---beyond the reach of PGD-based methods. This can lead to considerably improved RE as PWCF \textbf{(I)} can serve as a \emph{reliable supplement} to the state-of-the-art (SOTA) RE packages on $\ell_1$, $\ell_2$, and $\ell_\infty$ attacks, e.g. \texttt{AutoAttack}~\cite{croce2020reliable}, and \textbf{(II)} opens up the possibility of RE over a much wider range of attack models, e.g., general $\ell_p$ attacks with any $p > 0$ and more complicated ones such as perceptual attacks~\cite{laidlaw2021perceptual}. We remark that PWCF is also general enough to solve \cref{eq:min_distort}, but due to the limited preliminary results currently at hand, we leave it as future work.
\section{Technical background}
\label{sec:background}
\cref{eq:robust_loss} is often solved by the projected gradient descent (PGD)\footnote{It should be ``ascent" instead of ``descent" due to the maximization, but we follow the \texttt{AutoAttack} package.} method. The basic update reads $\mb x'_{new} = \mc P_{\Delta(\mb x)} (\mb x'_{old} + t \nabla \ell(\mb x'_{old}))$, where $\mc P_{\Delta(\mb x)}$ is the projection operator onto $\Delta(\mb x)$. When $\Delta(\mb x) = \{\mb x' \in [0, 1]^n: \norm{\mb x' - \mb x}_p \le \eps\}$ with $p = 1, \infty$, $\mc P_{\Delta(\mb x)}$ takes simple forms. For $p=2$, sequential projection onto the box and then the norm ball at least finds a feasible solution. Hence, PGD is feasible for these cases. For \emph{other choices of $p$} and \emph{general non-$\ell_p$ metrics $d$} where analytical projection is not so intuitive to derive, existing PGD based algorithms does not apply. For practical PGD methods, previous works have shown that the solution quality is sensitive to the tuning of multiple hyperparameters, e.g., step-size schedule and iteration budget~\cite{mosbach2018logit,CarliniEtAl2019Evaluating,croce2020reliable}. The SOTA PGD variants, APGD-CE and APGD-DLR, try to make the tuning automatic by combining a heuristic adaptive step-size schedule and momentum acceleration under fixed iteration budget~\cite{croce2020reliable}---both are built into the popular \texttt{AutoAttack} package\footnote{\url{https://github.com/fra31/auto-attack}}.

\subsection{\texttt{PyGRANSO} for constrained optimization}

In principle, as an instance of nonlinear optimization (NO) problems~\cite{Bertsekas2016Nonlinear}
\begin{align}  \label{eq:NO_form}
\min_{\mb x}\; g(\mb x) ~ \text{,} \quad
\st \; c_i(\mb x) \le 0 \; \forall\; i \in\mc I; \; h_j(\mb x) = 0 \; \forall\; j \in \mc E 
\end{align}
\cref{eq:robust_loss} can be solved by general-purpose NO solvers such as \texttt{Knitro}~\cite{pillo2006large}, \texttt{Ipopt}~\cite{wachter2006implementation}, and \texttt{GENO}~\cite{laue2019geno}. However, there are two caveats: (1) the above solvers only handle continuously differentiable objective and constraint functions, i.e., $g$, $c_i$'s, and $h_j$'s, but non-differentiable $g$, $c_i$'s, and $h_j$'s are common in \cref{eq:robust_loss}, e.g., when $d$ is the $\ell_1$ or $\ell_\infty$ distance, or $f_{\mb \theta}$ uses non-differentiable activations;  (2) they require analytical gradients of $g$, $c_i$'s, and $h_j$'s, which are impractical to derive when DNN models $f_{\mb \theta}$ are involved. 

\pygranso\footnote{\url{https://ncvx.org}}~\cite{BuyunLiangSun2021NCVX,curtis2017bfgs} is a recent \texttt{PyTorch}-port of the powerful MATLAB package \texttt{GRANSO}~\cite{curtis2017bfgs} which can handle general NO problems of form \cref{eq:NO_form} and potentially with non-differentiable $g$, $c_i$'s, and $h_j$'s. It only requires these functions to be \emph{almost everywhere differentiable}, which is satisfied by almost all forms of \cref{eq:robust_loss} proposed so far in the literature. \texttt{GRANSO} employs a quasi-Newton sequential quadratic programming (BFGS-SQP) to solve \cref{eq:NO_form}, and features a rigorous adaptive step-size rule via line search and a principled stopping criterion inspired by gradient sampling~\cite{burke2020gradient}. \pygranso\,equips \texttt{GRANSO} with auto-differentiation and GPU computing powered by \texttt{PyTorch}---crucial for deep learning problems. The stopping criterion is controlled by stationarity, total constraint violation, and optimization tolerance---all can be transparently controlled, but is typically unnecessary to tune. For the details of \texttt{PyGRANSO} package, please check: \url{https://arxiv.org/abs/2210.00973}.

\section{\pygranso\, with constraint folding as a generic solver for \cref{eq:robust_loss}}  \label{Sec:pygranso}

Though \pygranso\ can serve as a promising solver for \cref{eq:robust_loss} with general metric $d$, we find in practice that naive deployment can suffer from slow convergence, or low quality solutions due to numerical issues. Below, we introduce \pygranso~\textbf{W}ith \textbf{C}onstraint-\textbf{F}olding (PWCF), and other techniques that can substantially speed up the optimization process, and improve the solution quality.

\subsection{Reformulating $\ell_\infty$ constraint to avoid sparse subgradients}
\label{subsec: reformulate Linf constraint}
The BFGS-SQP algorithm inside \pygranso\, relies on the subgradients of the objective and the constraint functions to approximate the (inverse) Hessian and to compute the search direction. Hence, when the subgradients are sparse, updating all optimization variables may take many iterations, leading to slow convergence. For the $\ell_\infty$ metric, 
\begin{align} \label{eq:subgrad_linf}
    \partial_{\mb z}  \norm{\mb z}_\infty = \conv \{\mb e_k \sign(z_k): z_k =  \norm{\mb z}_\infty \; \forall\, k\}, 
\end{align} 
where $\mb e_k$'s are the standard basis vectors, $\conv$ denotes convex hull, and $\sign(z_k) = z_k/\abs{z_k}$ if $z_k \ne 0$, else $[-1, 1]$.
The subgradient in \cref{eq:subgrad_linf} contains no more than $n_k = \abs{\{k: z_k = \norm{\mb z}_\infty\}}$ nonzeros, and hence is sparse when $n_k$ is small. To avoid this issue, we propose a reformulation
\begin{equation}
    \label{eq: Linf to box}
    \norm{\mb x - \mb x'}_\infty \le \eps \Longleftrightarrow -\eps \mb 1 \le \mb x - \mb x' \le \eps \mb 1. 
\end{equation}

\subsection{Constraint-folding to reduce the number of constraints}
\label{subsec: folding}
The natural image constraint $\mb x' \in [0, 1]^n$ is a set of $n$ box constraints. The reformulation described in \cref{subsec: reformulate Linf constraint} introduces another $\Theta(n)$ box constraints. Although all these are just simple linear constraints, the $\Theta(n)$-growth is daunting: for natural images, $n$ is the number of pixels that can easily get into hundreds of thousands. Typical NO problems become more difficult the number of constraints grows, e.g., leading to slow convergence for numerical algorithms. 

To combat this, we introduce a folding technique that can reduce the number of constraints into a small constant. To see how this is possible, first note that any equality constraint $h(\mb x) = 0$ or inequality constraint $c(\mb x) \le 0$ can be reformulated as 
\begin{align} \label{eq:simple_constr_form}
h (\mb x) = 0 \Longleftrightarrow \abs{h(\mb x)} \le 0 ~ \text{,} \quad
c(\mb x) \le 0 \Longleftrightarrow \max\{c(\mb x), 0\} \le 0. 
\end{align} 
We can then fold them together as 
\begin{equation} 
    \label{eq:folded_constraint} 
    \mc F(|h(\mb x)|, \max\{c(\mb x), 0\}) \le 0, 
\end{equation}
where $\mc F: \R_+^{2} \mapsto \R_+$ ($\R_+ \doteq \{t: t \ge 0\}$) can be any function satisfying $\mc F(\mb z) = 0 \Longrightarrow \mb z = \mb 0$, e.g., any $\ell_p$ ($p \ge 1$) norm.

\begin{figure}[!htbp]
    \centering
    \includegraphics[width=0.95\textwidth]{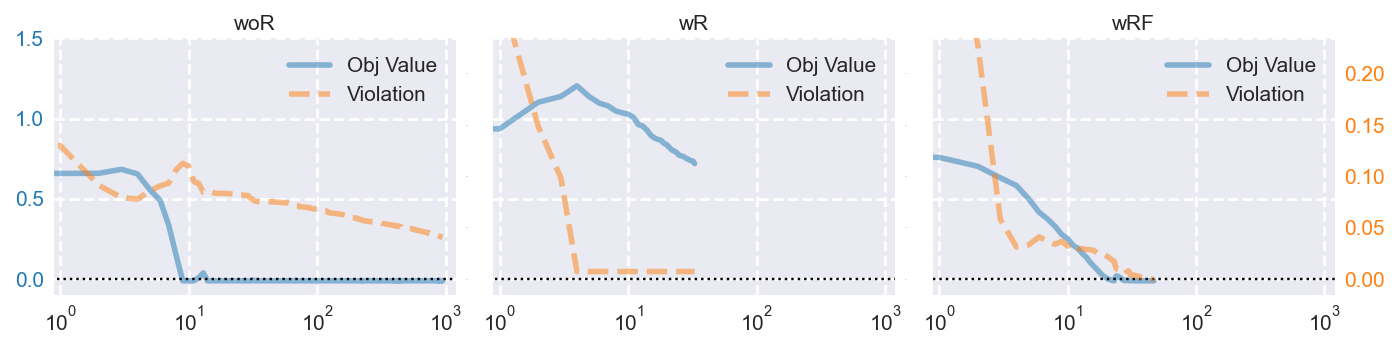}
    \vspace{-1em}
    \caption{Optimization trajectory of the \textbf{objective value} and \textbf{constraint violation} w.r.t iterations for an $\ell_\infty$ case on CIFAR-10 dataset. \textbf{woR}: using $\ell_\infty$ original form; \textbf{wR}: with reformulation but no folding; \textbf{wRF}: with reformulation and folding. Maximum time budget per curve: $600s$ (only wRF terminates before reaching this budget). Both \textbf{objective} and \textbf{violation} reaching $0$ indicates successful attack.} 
    \label{fig:benefit_reform_folding}
    \vspace{-1em}
\end{figure}
It is easy to verify the equivalence of \cref{eq:folded_constraint} and the original constraints in \cref{eq:simple_constr_form}. The folding technique can be used to a subset or all of the constraints; one can group and then fold constraints according to their physical meanings. We note that folding or aggregating constraints is not a new idea and has been popular in engineering design. For example, \cite{martins2005structural} uses $\ell_\infty$ folding and its log-sum-exponential approximation to deal with numerous design constraints. However, applying folding into NO problems in machine learning seems rare, potentially because producing non-differentiable constraint(s) due to the folding seems counterproductive. 

In our experiments, we use $\mc F = \norm{\cdot}_2$ to fold the $\Theta(n)$ box constraints from $\ell_{\infty}$ reformulation into a single constraint, enforce the $\mb x' \in [0, 1]^n$ constraints in $f_{\mb \theta}$ by direct clipping. \cref{fig:benefit_reform_folding} shows clearly that combining folding and reformulation can substantially speed up convergence and boost the solution quality for our algorithm.  

\subsection{Loss clipping when solving \cref{eq:robust_loss} with PWCF}
For \cref{eq:robust_loss} with the popular cross-entropy (CE) and margin losses, the objective value can easily dominate constraint violation during the maximization process. Since \pygranso\,tries to balance the objective value and constraint violation when making progress, it can persistently prioritize optimizing the objective over constraint satisfaction, resulting in very slow progress in finding a feasible solution. To resolve this numerical difficulty, we propose using clipped margin loss $\ell_{ML}$ with maximal value $0.01$, as any $\ell_{ML} \ge 0$ indicates a successful attack. For the same reason, we use clipped CE loss with maximal value at $10$ in PWCF\footnote{Attack success happens when the true logit output less than $1/K$ (assuming softmax normalization is applied), where $K$ is the number of classes. So the critical value is $-\log 1/K$, which is $< 10$ for $K \le e^{10}$, sufficient for typical RE datasets. }.

\section{Experiments and results: solving \cref{eq:robust_loss} with PWCF}
\subsection{PWCF offers competitive and complementary attack performance to \cref{eq:robust_loss}}
\label{subsec: max formulation with l1, l2, linf}

We take SOTA $\ell_1$-, $\ell_2$-, and $\ell_\infty$-adversarially trained models on CIFAR10\footnote{\url{https://github.com/locuslab/robust_union/tree/master/CIFAR10}}\footnote{\url{https://github.com/deepmind/deepmind-research/tree/master/adversarial_robustness}}, and an adversarially-trained model with respect to the LPIPS distance\footnote{See \cref{subsec: formulation with general lp norm} for details.} on ImageNet~\cite{laidlaw2021perceptual}\footnote{\url{https://github.com/cassidylaidlaw/perceptual-advex}}, to compare the attack performance by solving \cref{eq:robust_loss} between PWCF and the APGD\footnote{We implement the margin loss on top of \texttt{AutoAttack}.}~\cite{Croce2020ReliableEO} method from \texttt{AutoAttack} package. 
The attack radii $\eps$'s are set following the common practice of adversarial RE\footnote{E.g., \url{https://robustbench.github.io/} for Cifar10 $\ell_2$ and $\ell_\infty$; \url{https://github.com/locuslab/robust_union} for Cifar10 $\ell_1$; \cite{laidlaw2021perceptual} for ImageNet $\ell_2$ and $\ell_\infty$.}.

\begin{table}[!htbp]
\vspace{-1em}
\caption{\textbf{Comparison of our PWCF with SOTA attack methods on 
$\ell_{1}$-, $\ell_{2}$- and $\ell_{\infty}$- attacks.} For given pretrained models, we report the models' \textbf{clean} and \textbf{robust} accuracy---lower \textbf{robust} accuracy means more effective attacks. We test on both CE and margin loss for APGD and PWCF. Numbers are in $(\%)$. \textbf{Model - Attack} denotes the selection of the models and the type of the performed adversarial attacks and its $\eps$.
}
\vspace{-1em}


\label{tab: granso_l1_acc}
\begin{center}
\setlength{\tabcolsep}{1.0mm}{
\begin{tabular}{l l c c c c c c c c c c c}
{}
&{}
&{}
&{}
&\multicolumn{2}{c}{\small{\textbf{APGD}}}
&{}
&\multicolumn{2}{c}{\small{\textbf{PWCF(ours)}}}
&{}
&\small{\textbf{Square}}
&{}
&\small{\textbf{APGD}}
\\
\cline{5-6}\cline{8-9}\cline{11-11}
\vspace{-8pt}
\\
{\small{\textbf{Dataset}}}
&{\small{\textbf{Model - Attack}}}
& \small{\textbf{Clean}}
&{}
& \small{\textbf{CE}}
& \small{\textbf{M}}
& {}
& \small{\textbf{CE}}
& \small{\textbf{M}}
& {}
& \small{\textbf{M}}
&{}
&\small{\textbf{+PWCF}}
\\
\toprule
{\small{$\text{CIFAR}10$}}
&\small{{P\textsubscript{1}~\cite{maini2020adversarial} - \textsubscript{$\ell_{1}(12)$}}}
&{$73.3$}
&{}
&{$0.96$}
&\textcolor{red}{$0.00$}
&{}
&\textcolor{black}{$28.6$}
&\textcolor{red}{$0.00$}
&{}
&{$2.28$}
&{}
&\textbf{0.00}
\\
\cline{2-13}
\vspace{-6pt}
\\
{}
&\small{{WRN\textsubscript{-70-16}~\cite{gowal2020uncovering} - \textsubscript{$\ell_{2}(0.5)$}}}
&{$94.7$}
&{}
&{$81.8$}
&{$81.1$}
&{}
&{$81.8$}
&\textcolor{red}{$81.0$}
&{}
&{$87.9$}
&{}
&\textbf{80.8}
\\
\cline{2-13}
\vspace{-8pt}
\\
{}
&\small{{WRN\textsubscript{-70-16}~\cite{gowal2020uncovering} - \textsubscript{$\ell_{\infty}(0.03)$}}}
&{$90.8$}
&{}
&{$69.4$}
&\textcolor{red}{$68.0$}
&{}
&{$73.6$}
&{$72.8$}
&{}
&{$71.6$}
&{}
&\textbf{67.1}
\\
\midrule
\midrule
{\small{$\text{ImageNet}100$}}
&\small{{PAT\textsubscript{-Alex}~\cite{laidlaw2021perceptual} - \textsubscript{$\ell_{2}(4.7)$}}}
&{$75.0$}
&{}
&\textcolor{red}{$42.7$}
&{$44.0$}
&{}
&{$42.8$}
&{$44.5$}
&{}
&{$63.1$}
&{}
&\textbf{40.9}
\\
\cline{2-13}
\vspace{-8pt}
\\
{}
&\small{{PAT\textsubscript{-Alex}~\cite{laidlaw2021perceptual} - \textsubscript{$\ell_\infty(0.016)$}}}
&{$75.0$}
&{}
&\textcolor{red}{$48.0$}
&{$48.2$}
&{}
&{$56.6$}
&{$48.8$}
&{}
&{$59.9$}
&{}
&\textbf{45.2}
\\
\bottomrule
\end{tabular}
}
\end{center}
\vspace{-1em}
\end{table}

From \cref{tab: granso_l1_acc}, we can conclude that: (1) PWCF performs strongly and comparably to APGD on $\ell_1$, $\ell_2$ and $\ell_\infty$ attacks, especially using \emph{margin loss} as the objective; (2) PWCF is weak on $\ell_1$ and $\ell_\infty$ attacks using CE loss, likely due to the bad numerical scaling of the CE loss; (3) Combining all successful attack samples found by APGD and PWCF (APGD+PWCF) can further reduce the robust accuracy compared to any single APGD or PWCF attack---PWCF and APGD are complementary. Note that~\cite{CarliniEtAl2019Evaluating} also remarks that the diversity of solutions matters much more than the superiority of individual solvers, which is the reason why \texttt{AutoAttack} includes Square Attack--a zero-th order black-box attack method that does not perform strongly itself as shown in \cref{tab: granso_l1_acc}.

\subsection{PWCF works for general (almost everywhere) differentiable $\ell_p$ and non-$\ell_p$ distances}
\label{subsec: formulation with general lp norm}

As highlighted in \cref{sec:background}, a major limitation of the PGD based solvers is that they cannot handle distances other than $\ell_1$, $\ell_2$, and $\ell_\infty$\footnote{We do not consider $\ell_0$ in this paper as it is not a norm, but we acknowledge that \cite{croce2019sparse} targets at generating $\ell_0$ attacks using PGD-based method.}.
By contrast, PWCF stands out as a convenient choice for general distances. To show this, we apply PWCF to solve \cref{eq:robust_loss} with $\ell_{1.5}$ and $\ell_{8}$ distances. In addition, we also solve \cref{eq:robust_loss} with the LPIPS perceptual metric~\cite{laidlaw2021perceptual,zhang2018unreasonable}, i.e., perceptual attack (PAT) with 
\begin{align}
\label{Eq. LPIPS Constraint} 
d(\mb x, \mb x') \doteq || \phi(\mb x) - \phi(\mb x')||_{2} ~ \text{,}  \quad \quad \phi(\mb x) \doteq [~\wh{g}_{1}(\mb x), \dots, \wh{g}_{L}(\mb x)~]
\end{align}
where $\wh{g}_{1}(\mb x), \dots, \wh{g}_{L}(\mb x)$ are the vectorized intermediate feature maps from pretrained DNNs.  

\begin{table}[!tb]
\centering 
\vspace{-1em}
\caption{\textbf{Attack performance of PWCF with margin loss on general $\ell_p$ and non-$\ell_p$ metrics}. We report attack success rates (numbers are in $\%$). We test on $\ell_{1.5}$, $\ell_{8}$, and PAT; numbers on $\ell_1$, $\ell_2$, and $\ell_\infty$ are included for reference. Numbers below each rate in parenthesis are the perturbation radii.
}
\vspace{-1em}

\label{tab: granso_lp_norm}
\setlength{\tabcolsep}{1.0mm}{
\begin{tabular}{l c c c c c c c}
{}
&\multicolumn{3}{c}{\small{\textbf{Special $\ell_p$}}}
&{}
&\multicolumn{3}{c}{\small{\textbf{General $\ell_p$}}}
\\
\cline{2-4}\cline{6-8}
\vspace{-8pt}
\\
\small{\textbf{Model}}
&\small{$\ell_1$}
&\small{$\ell_2$}
&\small{$\ell_\infty$}
&{}
&\small{$\ell_{1.5}$}
&\small{$\ell_8$}
&\small{PAT}
\\
\toprule
\small{Clean}
&{$100$}
&{$100$}
&{$100$}
&{}
&{$100$}
&{$100$}
&{$100$}
\\

&{{\footnotesize $(2400)$}}
&{{\footnotesize $(6.09)$}}
&{{\footnotesize $(0.01569)$}}
&{}
&{{\footnotesize $(44.40)$}}
&{{\footnotesize $(0.07)$}}
&{{\footnotesize $(0.5)$}}
\\
\small{PAT}
&{$49.7$}
&{$40.7$}
&{$35.2$}
&{}
&{$100$}
&{$100$}
&{$100$}
\\

&{{\footnotesize $(2400)$}}
&{{\footnotesize $(4.7)$}}
&{{\footnotesize $(0.017)$}}
&{}
&{{\footnotesize $(443.98)$}}
&{{\footnotesize $(0.70)$}}
&{{\footnotesize $(0.5)$}}
\\
\bottomrule
\end{tabular}
}
\vspace{-1em}
\end{table}

PWCF handles them seamlessly, as shown in 
\cref{tab: granso_lp_norm}. Here we do not strive to set the most reasonable perturbation radii, especially for $\ell_{1.5}$ and $\ell_8$ that have not been tested before, and hence we also do not stress the attack rates. Our point is that \emph{PWCF is able to handle these general $\ell_p$ distances}. \cref{tab: granso_pat_compare} further summarizes the details of performing the perceptual attack with $\eps=0.5$. Existing methods to compare are Perceptual Projected Gradient Descent (PPGD), Lagrangian perceptual attack (LPA) and its variant fast Lagrangian perceptual attack (Fast-LPA) methods, all developed in~\cite{laidlaw2021perceptual}, based on iterative linearization and projection (PPGD), or penalty method (LPA, Fast-LPA) respectively. 
In addition to the objective values and attack success rates, we also report their chances of finding infeasible solutions. As observed in \cref{tab: granso_pat_compare}, our PWCF is the clear winner.

\begin{table}[!tb]
\caption{\textbf{Performance comparison of different methods solving \textbf{PAT} with the clipped CE and margin (M) loss}. \textbf{Viol.} reports the ratio of final solutions that violate constraints. \textbf{Succ.} is the ratio of all \emph{feasible successful attacks} divided by \emph{total number of samples}. The model we test is \texttt{pat\_alexnet\_0.5}~\cite{laidlaw2021perceptual}. Evaluation is performed on ImageNet-100 dataset.}
\vspace{-1em}

\label{tab: granso_pat_compare}
\begin{center}
\setlength{\tabcolsep}{1.0mm}{
\begin{tabular}{l c c c c c}
{}
&\multicolumn{2}{c}{\small{\textbf{CE Objective}}}
&{}
&\multicolumn{2}{c}{\small{\textbf{Margin Objective}}}
\\
\cline{2-3}\cline{5-6}
\vspace{-8pt}
\\
\small{\textbf{Method}}
&\small{\textbf{Viol. ($\%$) $\downarrow$}}
&\small{\textbf{Succ. ($\%$) $\uparrow$}}
&{}
&\small{\textbf{Viol. ($\%$) $\downarrow$}}
&\small{\textbf{Succ. ($\%$) $\uparrow$}}
\\
\toprule
\small{Fast-LPA}
&{$73.8$}
&\textcolor{black}{$3.54$}
&{}
&{$41.6$}
&{$56.8$}
\\
\small{LPA}
&{\textbf{0.00}}
&{$80.5$}
&{}
&{\textbf{0.00}}
&\textcolor{black}{$97.0$}
\\
\small{PPGD}
&{$5.44$}
&{$25.5$}
&{}
&{\textbf{0.00}}
&{$38.5$}
\\
\midrule
\small{PWCF}
&{$0.62$}
&\textcolor{red}{$93.6$}
&{}
&{\textbf{0.00}}
&\textcolor{red}{$100$}
\\
\bottomrule
\end{tabular}
}
\end{center}
\vspace{-2em}
\end{table}

\section{Conclusion}
\label{sec:conclusion} 

In this paper, we propose PWCF to solve the maximization problem \cref{eq:robust_loss} in robustness evaluations, blending the SOTA constrained optimization solver \texttt{PyGRANSO} with constraint folding and other tweaks. Our experimental results show that 1) PWCF can provide competitive and complementary performance compared with the SOTA methods on $\ell_1$, $\ell_2$, and $\ell_\infty$ attacks; 2) PWCF can deal with general attack models such as $\ell_{p}$ with $p \ge 1$ and perceptual attacks, which are beyond the reach of existing PGD-based methods; 3) PWCF involves little to zero parameter-tuning and obtains reliable solutions based on a principled stopping criterion. Our preliminary experiments also show that the proposed PWCF is general enough to solve \cref{eq:min_distort} with good quality, which we will present in forthcoming papers.

\vfill\pagebreak
\bibliography{reference,egbib,theory}

\end{document}